\documentclass[letterpaper, 10 pt, conference]{ieeeconf}  
\IEEEoverridecommandlockouts
\usepackage{graphicx}
\usepackage{tabularx}
\usepackage{booktabs}
\usepackage{nicefrac}
\usepackage{multirow}
\usepackage{capt-of}
\usepackage{gensymb}
\usepackage{makecell}
\usepackage{xcolor}
\usepackage{tikz}
\usepackage[hidelinks]{hyperref}
\graphicspath{ {images/} }

\newcommand\copyrighttext{%
  \footnotesize \textcopyright 2022 IEEE. Personal use of this material is permitted. Permission from IEEE must be obtained for all other uses, in any current or future media, including reprinting/republishing this material for advertising or promotional purposes, creating new collective works, for resale or redistribution to servers or lists, or reuse of any copyrighted component of this work in other works.}
\newcommand\copyrightnotice{%
\begin{tikzpicture}[remember picture,overlay]
\node[anchor=south,yshift=10pt] at (current page.south) {\fbox{\parbox{\dimexpr\textwidth-\fboxsep-\fboxrule\relax}{\copyrighttext}}};
\end{tikzpicture}%
}

\title{\LARGE \bf
Design of Spiral-Cable Forearm Exoskeleton to Assist Supination for Hemiparetic Stroke Subjects
}

\author{Ava Chen$^{1}$, Lauren Winterbottom$^{2}$, Katherine O'Reilly$^{1}$, Sangwoo Park${^1}$, Dawn Nilsen$^{2,3}$,\\ Joel Stein$^{2,3}$, and Matei Ciocarlie$^{1,3}$%
\thanks{This work was supported in part by the National Institute of Neurological Disorders and Stroke under grant R01NS115652}%
\thanks{$^{1}$Department of Mechanical Engineering, Columbia University, New York, NY 10027, USA.
        {\tt\small \{ava.chen, k.oreilly, sp3287, matei.ciocarlie\}@columbia.edu}}%
\thanks{$^{2}$Department of Rehabilitation and Regenerative Medicine, Columbia University, New York, NY 10032, USA.
        {\tt\small \{lbw2136, dmn12, js1165\}@cumc.columbia.edu}}%
\thanks{$^{3}$Co-Principal Investigators}
}

\begin{document}

\maketitle
\copyrightnotice
\thispagestyle{empty}
\pagestyle{empty}
\begin{abstract}
We present the development of a cable-based passive forearm exoskeleton that is designed to assist supination for hemiparetic stroke survivors. Our device uniquely provides torque sufficient for counteracting spasticity within a below-elbow apparatus. The mechanism consists of a spiral single-tendon routing embedded in a rigid forearm brace and terminated at the hand and upper-forearm. A spool with an internal releasable-ratchet mechanism allows the user to manually retract the tendon and rotate the hand to counteract involuntary pronation synergies due to stroke. We characterize the mechanism with benchtop testing and five healthy subjects, and perform a preliminary assessment of the exoskeleton with a single chronic stroke subject having minimal supination ability. The mechanism can be integrated into an existing active hand-opening orthosis to enable supination support during grasping tasks, and also allows for a future actuated supination strategy.
\end{abstract}

\section{Introduction}
Wearable assistive devices have established themselves as promising tools for supplementing an impaired hand's function and encouraging its use in everyday life \cite{vanommeren2018}. For example, we \mbox{\cite{park2020,park2019}} and other groups \cite{parker2020} have introduced a number of robotic devices to assist hemiparetic stroke survivors with grasping tasks. However, assisting only grasping when a user lacks active forearm rotation severely limits the functional workspace of the hand and makes many daily tasks difficult or \mbox{impossible \cite{nam2019}.} We informally observed this phenomenon in our own work with chronic stroke survivors, where several subjects would attempt to grasp cups or bottles from a pronated hand position and had difficulty supinating the forearm enough to reliably grasp and use these objects. Flexor synergies from squeezing objects in the hand or lifting the arm further exacerbated unwanted pronation. These patient-driven observations motivated the development of a lightweight exoskeleton specifically designed to assist supination that featured a small footprint feasible for integration in existing below-elbow grasping exoskeletons.

\begin{figure}[t]
    \centering
    \includegraphics[width=0.9\columnwidth]{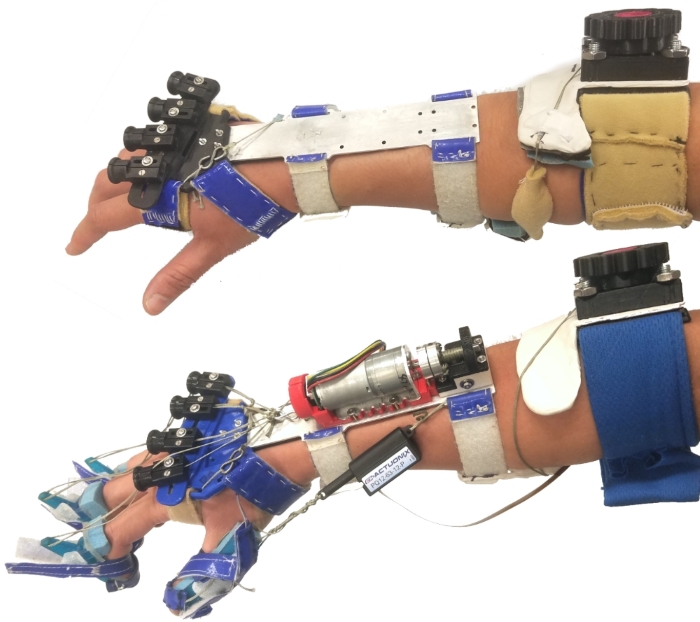}
    \vspace{-.1cm}
    \caption{Supination exoskeleton worn as a standalone device (Top) and as an integrated supination and hand-opening orthosis to assist grasping (Bottom).}
    \label{fig:robot}
    \vspace{-.6cm}
\end{figure}

Although many instrumented benchtop devices and workstation-mounted exoskeletons have been developed to provide rehabilitative therapy in supination \cite{wang2021, panny2020, lambercy2011, pehlivan2011}, there are few devices that provide rotational assistance in a fully-wearable form. SpringWear, developed by Chen and \mbox{Lum \cite{chenj2018}}, uses a shoulder-anchored forearm brace with an elbow linkage mechanism. Gasser \mbox{et al.} \cite{gasser2020} implemented forearm  rotation adjustment with a lockable ball-detent mechanism as part of a rigid elbow-joint module within a hand-arm orthosis. Myomo's MyoPro Motion G is a commercial device that combines these concepts by using a ball-detent system within a C-shaped cuff at the wrist \cite{peters2017}. But these all require precise positioning about several joints; in particular, obstruction of shoulder and elbow movement for many stroke survivors encourages maladaptive compensation from the trunk to perform reaching movements and manage hand pose \cite{rb2021}. Commercial spiral-supinator elastic bands anchor around the bicep and thumb to provide constant passive resistance against pronation, with minimal constraint to elbow flexion. However, elastic bands couple supination with arm extension and cannot be easily adjusted when worn underneath a hand orthosis. Soft-actuator versions of the spiral \mbox{concept \cite{parksh2019},} including one contained within a below-elbow \mbox{sleeve \cite{realmuto2019},} use inflatable bladders enveloping the entire forearm that similarly make integration difficult.

This article describes the design of a passive forearm exoskeleton (Fig. \ref{fig:robot}) and embedded spiral cable route (Fig. \ref{fig:tendonroute}) that converts torque about a hand-turned dial to supination torque about the forearm, enabling adjustment of forearm rotation angle. The dial incorporates a ratcheting mechanism to lock the device at the desired position; tension in the cable counteracts spasticity and stroke-imposed flexor synergies. Our device is specifically intended for individuals with unidirectional deficit in supination; thus, we focus on providing unidirectional assistance and include a release mechanism to re-enable unrestricted, body-powered pronation.

To our knowledge, we are the first to present a physical wearable device featuring a cable-based mechanism for adjusting forearm rotation. This implementation is uniquely able to leave the elbow unencumbered, instead using differential rotation (Fig. \ref{fig:beambend}) along the length of the forearm to apply  torque. Our device is designed to assist repositioning from a fully-pronated to neutral hand position, doing so with an easily adjustable, unobtrusive package that can be integrated in an existing robotic hand orthosis. We characterize device angle and force--torque metrics with testing on able-bodied subjects, then conduct a preliminary feasibility test of the device's potential to assist maintaining a neutral arm position in a case study with one chronic stroke subject having minimal supination ability. 

\section{Supination Forearm Mechanism}

The robotic hand exoskeleton shown in Fig. \ref{fig:robot} consists of two independent, integrated mechanisms mounted to a rigid, aluminum forearm splint: a two-actuator finger-thumb extension device connected to 3D-printed components strapped to the distal links of the digits, and a manually-adjustable supination device strapped around the hand and forearm (shown alone in Fig. \ref{fig:robot}, Top). The design of the digit extension device used in this study was previously described in \cite{park2020,park2019} whereas the supination exoskeleton has not been previously described. For our hemiparetic population of interest, lack of supination function is usually accompanied by hand impairment; although the supination exoskeleton could be used as a standalone device, its intended usage is in conjunction with passive or robotic hand orthoses that assist grasping. We designed the supination exoskeleton to be compatible with typical hand orthoses that attach to the body with straps encircling the forearm and have their actuation apparatus on the dorsal side.

\subsection{Spiral-cable Actuation Principle}
    
\begin{figure}[t]
    \centering
    \vspace{.1cm}
    \includegraphics[width=0.92\columnwidth]{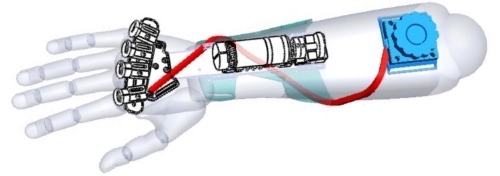}
    \vspace{-.1cm}
    \caption{Diagram highlighting spiral exotendon-routing concept. The tendon (red) is anchored to an orthosis on the dorsal side of the hand, and is guided in a spiral path around the arm via a sheath embedded in a thermoplastic brace (cyan) strapped to the forearm. This cable can be retracted within a hand-turned winch (blue) mounted at the base of the forearm; cable tension imposes supination torques about the wrist.}
    \label{fig:tendonroute}
    \vspace{.2cm}
    \includegraphics[width=0.9\columnwidth]{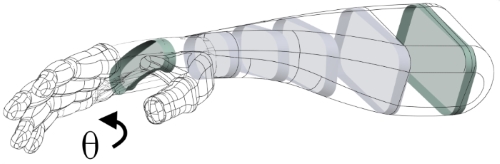}
    \vspace{-.1cm}
    \caption{Supination is considered in this paper as a distributed-rotation motion along the forearm. The proposed spiral cable mechanism generates relative torque between the hand and proximal region of the forearm (sections highlighted green). Use of the forearm itself as a reference plane enables the device to be wholly contained below the elbow.}
    \label{fig:beambend}
    \vspace{.2cm}
    \includegraphics[width=0.85\columnwidth]{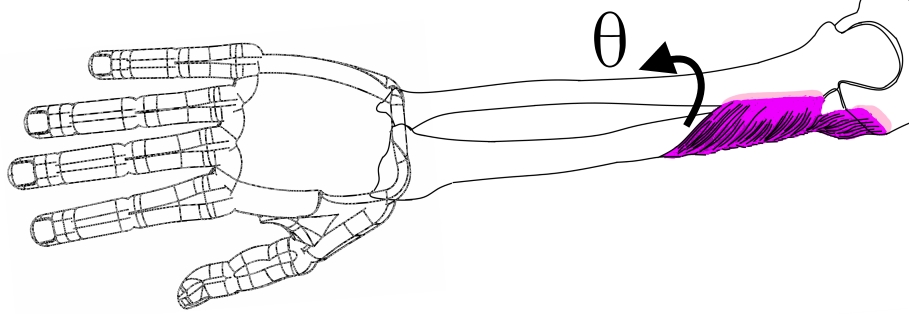}
    \vspace{-.1cm}
    \caption{Biomechanics of the supinator muscle: the supinator (purple) anchors to the ulna and humerus, and encircles the radius like a sling before attaching to the bone. Its contraction rotates the radius against the ulna.}
    \label{fig:bones}
    \vspace{-.5cm}
\end{figure}

Our design is inspired by the supinator (\textit{supinator brevis}) muscle, highlighted in purple in Fig. \ref{fig:bones}, which occupies the posterior compartment of the forearm and curves around the upper third of the radius to connect to the ulna. Muscle contraction pulls the radius laterally, turning the arm about its longitudinal axis. The exoskeleton likewise rotates the wrist in supination by pulling a cable wrapped around the forearm, highlighted in red in \mbox{Fig. \ref{fig:tendonroute}.} The cable is anchored at the hand and at the posterior area of the forearm. This design relies on the hand-wrist component of the exoskeleton having sufficient rotation relative to the proximal region of the forearm. Because there is a lack of literature on torque requirements needed to resist pronation and spasticity in a hemiparetic stroke population, we set an initial goal of achieving 0.4--0.5 Nm supination torque (approx. 12--15 N cable tension) and $60^\circ$ range of motion (ROM) based on reported actuator performance of related devices \cite{chenj2018,parksh2019}.

Because the device rigidly splints the wrist, the hand-arm system acts similarly to a cantilever beam in torsion. When the arm is subjected to an axial moment, each cross-section twists about its torsional center. The angle of twist decays as axial distance from the applied torque increases, generating a distributed rotation as illustrated in Fig. \ref{fig:beambend}. We maximize effective distance by placing the cable's distal termination point on the orthosis at the base of the index finger.

Any external twist will impose substantial shear and warping stresses on the arm. To prevent potential discomfort or torque losses from cable constriction, the exoskeletal housing for the cable must be sufficiently rigid in compression to prevent transmitting undesired pressures to the soft tissue. Additionally, the exoskeleton must prevent collapse of the spiral routing when tension is applied to the cable. We accomplish these criteria by routing the cable through a sheath embedded within a rigid thermoplastic brace.

The supination exotendon's proximal termination point is mounted to a mechanism strapped to the bulk soft tissue of the forearm, which is a compliant structure that will deform under cable tension. We note that our design requirement for sufficient relative rotation between hand and forearm can accommodate twisting losses due to soft tissue deformation so long as the proximal anchor does not slip against the skin.

\subsection{Forearm Brace}

Practical considerations for the design of an exoskeleton to house a cable tightly wrapped around the arm include preventing the cable from constricting around soft tissue, mitigating risk of the cable chafing against skin, and aiding ease of donning/doffing while transmitting sufficient torques.

\begin{figure}[t!]
    \centering
    \vspace{.1cm}
    \includegraphics[width=\columnwidth]{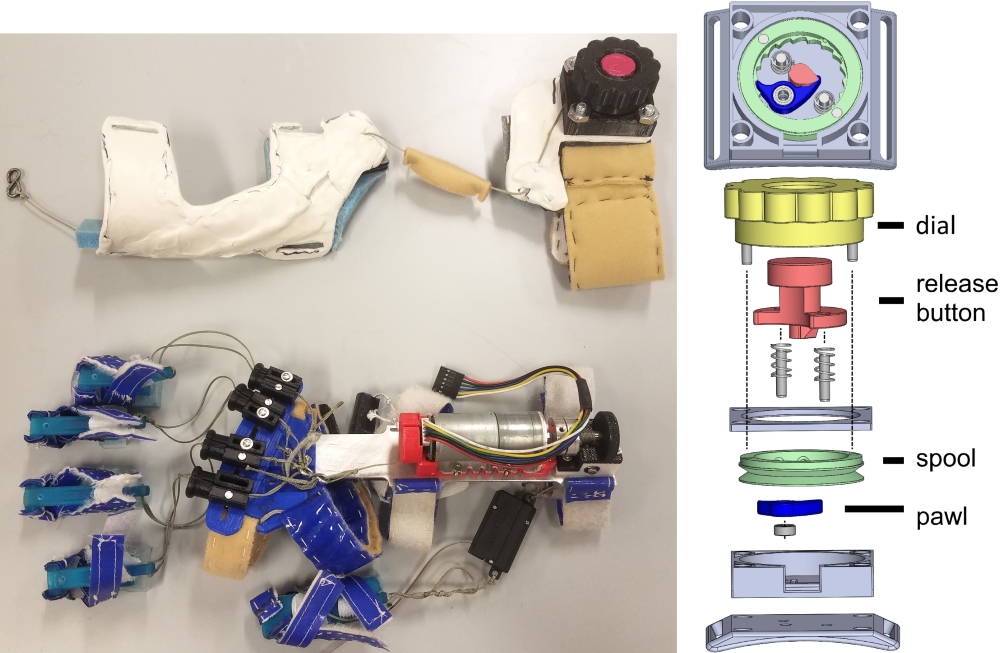}
    \caption{Left: the supination brace and hand orthosis used for this study. Right: assembly and internals of hand-turned cable adjustment mechanism. }
    \label{fig:splitrobot}
    \vspace{-.4cm}
\end{figure}

The proposed solution is a two-part brace, for which one part is a mid-forearm open-C brace secured with fabric straps shared with the hand-opening orthosis and the other part is an armband strapped around the proximal forearm (Fig. \ref{fig:splitrobot}, Left). The brace geometry is designed to unclasp with the same motions used to doff the hand-opening brace, adding only one operation to unstrap the proximal armband. The supination cable exits a spool mounted dorsally on the armband, is guided circumferentially along the forearm over a plastic skid plate, then crosses an elevated gap above the forearm to enter the spiral teflon sheath embedded within the distal brace. Both components of the supination brace are made from self-bonding thermoplastic molded to match the shape of the forearm. Altogether, the supination module has a mass of 161.1 g, for a total mass of \mbox{445.7 g} when integrated with the hand-opening orthosis.

Soft high-density foam padding is positioned in the interior of the brace and underneath the armband's skid plate to elevate the exoskeleton where the cable crosses the open gap between components, mitigating risk of the plastic plates pinching against the skin when the cable is retracted. A fabric pad attached around the cable at this gap serves to further minimize risk of skin pinching or chafing, although during normal usage the cable does not contact the skin even at full retraction. We also use foam padding to cushion the bony prominences at the wrist, so that we can more tightly secure the brace around the arm.

The fabric strap securing the armband to the forearm includes a layer of high-friction Dycem fabric sewn on the underside, such that the device moves with the skin instead of slipping. The armband is positioned on the user at the posterior compartment immediately below the elbow, but exact positioning is not required for function of the device.

\begin{figure*}[t]
    \centering
    \vspace{.16cm}
    \includegraphics[width=.99\textwidth]{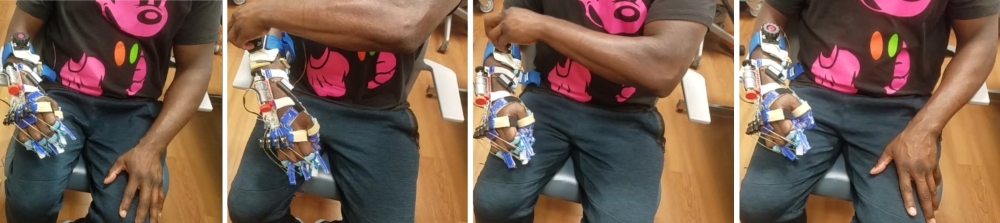}
    \vspace{-.2cm}
    \caption{Image series of stroke subject demonstrating usage of cable adjustment mechanism. From left to right: (1) subject wearing device in unlocked state; (2) reaching to turn adjustment dial; (3) tightening mechanism to supinate arm; (4) device fully-engaged, with arm fixed in a near-neutral orientation.}
    \label{fig:patientseries}
    \vspace{-.4cm}
\end{figure*}

Cable retraction is accomplished by rotating a dial attached to a ratchet assembly (Fig. \ref{fig:splitrobot}, Right). Twisting a dial located at the top of the device directly turns the spool to wind the cable and rotate the arm, as demonstrated in \mbox{Fig. \ref{fig:patientseries}.} The internal gear mechanism enables one-way progress capture and prevents backward rotation, allowing the user to set and hold a position to concentrate on grasping without having to maintain forearm exertion. The user can release the cable by pressing a button at the center of the dial to disengage the ratchet mechanism, allowing the arm to freely pronate. This design is similar to many commercial dial fasteners, but readily allows for the addition of a motor in future prototypes.

\section{Exoskeleton Validation}
For initial validation of the design, we characterized how our device withstands applied forces from the cable transmission and its ability to help a user to resist pronation in a series of experiments with a set of five able-bodied subjects and with a single stroke subject having minimal volitional supination. Testing occurred at Columbia University's Morningside Campus (able-bodied) and at NewYork-Presbyterian Hospital (stroke) in accordance with the protocol \mbox{(IRB-AAAS8104)} approved by the Columbia University Medical Center Institutional Review Board. Experiments with the stroke subject were performed under supervision of an occupational therapist.

Video clips of subjects performing experiments can be found at \mbox{\url{https://youtu.be/stlozxxWViA}}

\subsection{Benchtop Mechanical Tests}
We first characterized the ratcheting winch assembly in isolation in order to test the locking mechanism. An experimenter used a handheld force gauge (Omega DFG31-200) to apply forces against the mechanism, which was fixed in a vise. We measured compression forces required to release the mechanism when under tension, and conducted tensile testing to determine ultimate strength of the ratchet system.

To simulate unlocking the device while under tension, we tightened the dial to approximately \mbox{15--20 N} cable tension. We then measured the button force required to release the cable. Compression force after four trials was \mbox{(8.3 $\pm$ 1.5) N.}

We determined ultimate holding strength by progressively increasing cable tension until device failure. We then disassembled the mechanism to determine cause. Ultimate tensile force after three trials was \mbox{(115.3 $\pm$ 4.0) N}, and in all cases failure was due to fracture or material wear at the pawl.

\subsection{Able-Bodied Characterization}

Because evaluating feasibility requires studying device operation in conjunction with body-powered movements, we obtained the relationship between cable pulling force and forearm angles in a resistive task with five able-bodied subjects. We note that a healthy population still presents wide variance in hand-arm sizes and strengths, muscle tone, skin tightness, joint flexibility, and user comfort levels. We use this population to test the distributed-rotation concept given the inherent compliance of soft tissue, without the additional complexity from spasticity or impaired motor synergies.

Subjects were instructed to position the hand at a neutral $0^\circ$ angle defined throughout this paper as with the thumb oriented vertically), then twisted the dial to tighten the cable mechanism until snug. Subjects applied a user-defined ``moderate amount of effort" to pronate against device resistance, then returned to the neutral position after the device prevented further movement. This task was repeated four more times without adjusting the mechanism. After each set of five trials, subjects were asked if they experienced any discomfort and were allowed a short break to retighten the dial. Three runs of the task were performed for a total of 15 repetitions per subject. Throughout the experiment, subjects were instructed to hold the elbow in-place to minimize shoulder activation. Cable tension was recorded using a load cell (Futek LSB200-FSH00097) that was mounted inline with the mechanism, and end-effector angle was recorded using a handheld two-axis gyroscope \mbox{(STMicroelectronics L2G2IS)} with inclinometer display \mbox{(Syleo Apps, ``Simple Inclinometer").} Sensor data were synced over video \mbox{(30 fps),} for an overall sampling rate of 10 Hz. Five trials were excluded from analysis due to camera occlusion.

\begin{figure}[t]
    \centering
    \vspace{0.1cm}
    \includegraphics[width=0.95\columnwidth]{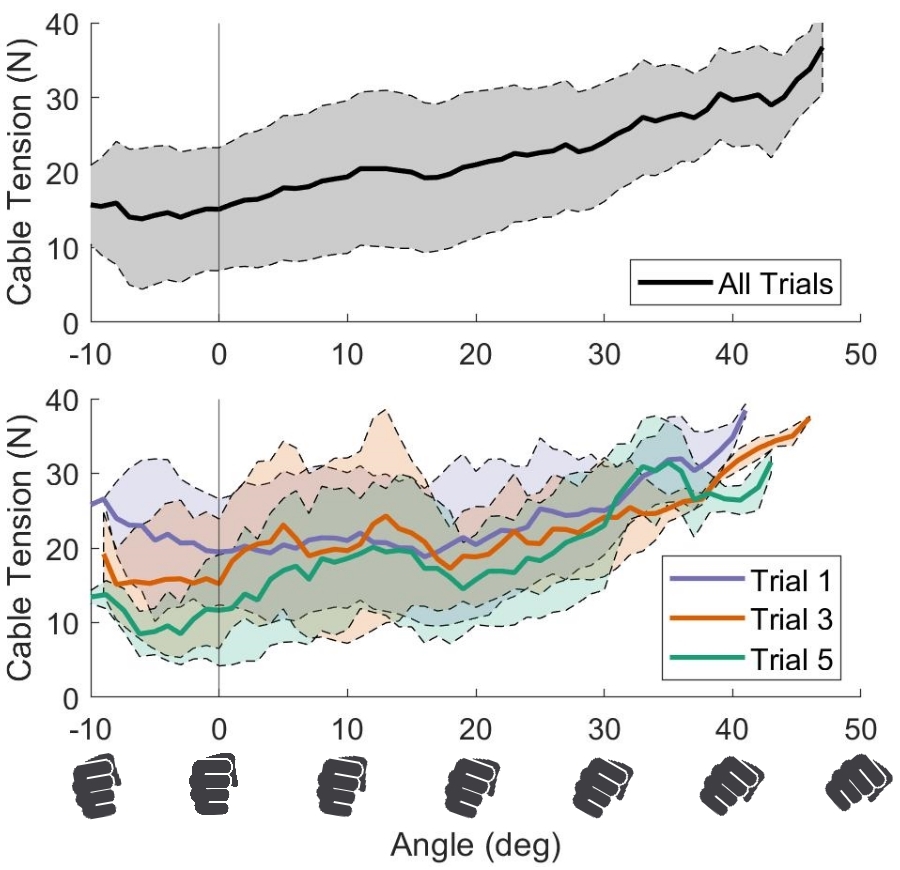}
    \vspace{-0.3cm}
    \caption{Top: able-bodied resistive loading results reported as mean and standard deviation aggregated across all five subjects and trials \mbox{(n = 70)}. Bottom: aggregated results for Trials 1, 3, and 5 \mbox{(n = 15, 15, 11)} depicting slight degradation of device performance over repeated cyclic strain. \\(Trials 2 and 4 are not shown to reduce figure clutter.)}
    \label{fig:able}
    \vspace{-0.5cm}
\end{figure}

Results of the able-bodied characterization are shown in Fig. \ref{fig:able}. We found that the proximal forearm anchor was generally able to resist pronation torques, as indicated by the increasing tension with hand angle. The effective rotational stiffness of the device varied between individuals, as we allowed users to self-select cable tightness and both magnitude and rate of applied pronation torque. For all subjects, compliance in the exoskeleton's straps and the change in forearm shape as the user strains against the device eventually allowed the arm to rotate within the brace to bypass the cable mechanism. This incremental slippage can be seen in the slight loss of performance over repeated cycles before the user retightens the cable. However, the device was able to meet the expected 15 N necessary to counteract spasticity throughout the ROM even accounting for device slippage. Our exoskeleton achieves comparable able-bodied ROM and torque performance to data reported by Realmuto and Sanger \cite{realmuto2019} while requiring a smaller device footprint. Subjects reported zero instances of discomfort during testing. 

\subsection{Feasibility Testing with a Hemiparetic Stroke Subject}

We evaluated the device on a single subject with right-sided hemiparesis and moderate spasticity in hand and arm; specifically, having scores of 2 for finger\slash elbow flexors and 1+ for forearm pronators on the Modified Ashworth Scale. Excessive spasticity disallowed the arm from being passively supinated past a neutral position. The subject was able to actively rotate the forearm $15^\circ$ out of a fully-pronated position \mbox{(Table \ref{tab:goniometer}),} and had sufficient arm strength to lift the hand to touch the face or extend the elbow. This subject was a participant in our previous studies with the hand-opening orthosis; however, involuntary pronation when lifting the arm prevented the subject from completing many of the functional tasks that required grasping cylindrical objects with a neutral hand pose. We first determined the subject's forearm pronation--supination (PS) ROM, validating our inclinometer method with standardized goniometer measurements. We then qualitatively evaluated the subject's functional ability to grasp and maneuver cylindrical objects with and without exoskeletal assistance. Our final set of tasks evaluated the device's ability to assist maintaining the hand in a neutral orientation when reaching. Because effects of spasticity vary daily, the tasks were conducted over the course of three 45-minute sessions, spaced three weeks apart. The subject reported zero instances of discomfort during the sessions.

\subsubsection{Forearm PS, elbow bent}
To study forearm rotation in isolation, we instructed the subject to sit with their shoulder adducted and elbow flexed at a right angle. With the inclinometer placed in their hand and with the arm resting on the chair's armrest, the subject was instructed to slightly lift the arm above the armrest (to prevent assistance from the chair) while attempting to keep the inclinometer in a vertical position. The subject performed this task without wearing the exoskeleton (baseline condition), while wearing the exoskeleton but without engaging the supination mechanism, and after fully retracting the cable mechanism (assisted condition). Experimenters manually rotated the subject's hand to measure passive ROM. To measure active pronation, we asked the subject to pronate as much as comfortably possible, then return to the baseline orientation. This task mimics the standard clinical procedure for measuring forearm PS ROM; we validated inclinometer readings by also taking goniometer measurements, which all agreed within 5$^\circ$. Results are shown in Table \ref{tab:goniometer}. With assistance, the subject was able to achieve a more-neutral position, but remained slightly pronated.

\begin{table}[h]
    \vspace{0.1cm}
    \centering
    \caption{Forearm PS ROM, Elbow Bent. Inclinometer Angular Distances from 0$^\circ$ Neutral Starting Position are reported.}
    \vspace{-.15cm}
    \begin{tabular}{cc@{---}ccr@{---}l}
    \toprule[1.5pt]
    Measurement (deg) & \multicolumn{2}{c}{Baseline} & & \multicolumn{2}{l}{With Device}\\
    \midrule
        Passive-ROM pronation & $0$\hspace{1pt} & $60$ & & --- & --- \\
        Passive-ROM supination & $0 $ & \hspace{3pt}$-35$ & & --- & --- \\
        Total Active ROM & $45$\hspace{3pt} & $60 $& & $25$\hspace{3pt} & \hspace{3pt}$50$\\
    \bottomrule
    \end{tabular}
    \label{tab:goniometer}
    \end{table}

\subsubsection{Hand Orientation, elbow extended}
The following set of reaching tasks focused on tracking total orientation of the hand instead of strictly measuring forearm PS rotation in isolation, since end-effector pose is most relevant for functional grasping. We note that hand orientation involves forearm PS as well as shoulder rotation, which we do not assist; however, we are interested in studying whether assisting forearm PS alone can enhance arm function. Overall, we expect the hand to pronate further when the subject reaches forward due to flexor-pronator synergies with the shoulder, along with forearm pronation. However, we attempted to limit effects of shoulder movement during these tasks by asking the subject to consciously keep their elbow tucked close to the body when performing the forward reach. This reduced habitual abduction and internal rotation of the shoulder.

\begin{figure}[t]
    \centering
    \vspace{.2cm}
    \includegraphics[width=0.8\columnwidth]{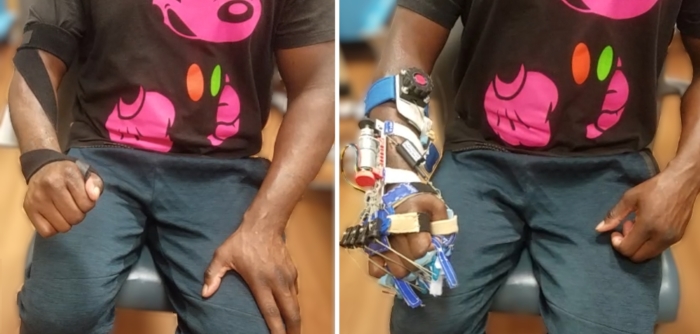}
    \caption{Stroke subject reaching forward while wearing an elastic supinator band (Left) and while wearing the proposed device (Right). Our device assists supination independent of arm extension.}
    \label{fig:mckiesplint}
    \vspace{.4cm}
    \includegraphics[width=0.8\columnwidth]{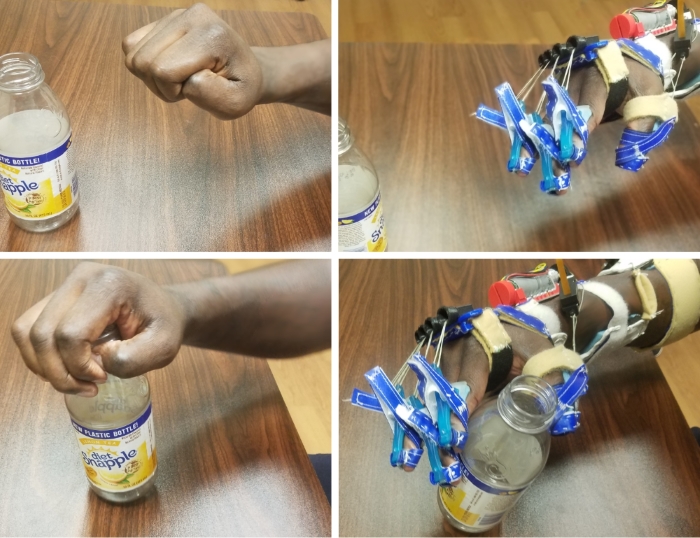}
    \caption{Stroke subject reaching for (Top) and grasping (Bottom) a bottle. The device enables a functional grasp for lifting the bottle to drink.}
    \label{fig:bottle}
    \vspace{-.5cm}
\end{figure}

\begin{figure*}[t]
    \centering
    \vspace{.3cm}
    \begin{minipage}[c]{.45\textwidth}
    \centering
    \includegraphics[width=\linewidth]{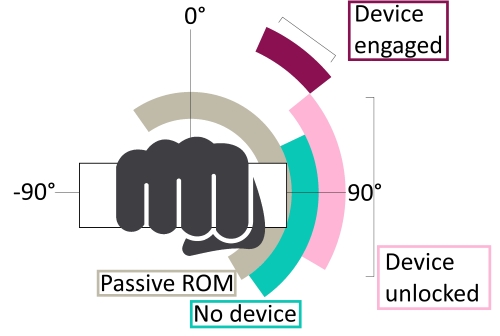}
    \end{minipage}
    \qquad
    \begin{minipage}[c]{.4\textwidth}
    \centering
    \vspace{-.1cm}
    \includegraphics[width=.9\linewidth]{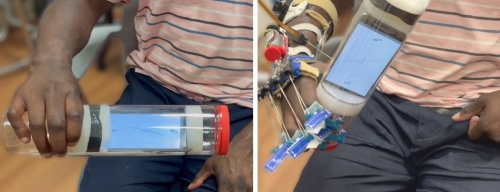}
    \vspace{-.2cm}
    \captionof{table}{Hand-Orientation Range When attempting to maintain Neutral 0$^\circ$ Position, Elbow Extended}
    \footnotesize
    \vspace{-.2cm}
    \begin{tabular}[b]{c@{\hspace{2.5em}}r@{ ---- }l}
        \toprule[1.5pt]
        Condition & \multicolumn{2}{c}{Angle Extent (deg)}\\
        \midrule
        No Device & \hspace{2em}$55$ & $150$ \\
        Device Disengaged & \hspace{2em}$45$ & $125$ \\
        Device Engaged & \hspace{2em}$15$ & $45$\\
        \bottomrule
    \end{tabular}
    \label{tab:angles}
    \end{minipage}
    \caption{Table \ref{tab:angles} and Left: total extent of inclinometer readings when stroke subject attempts to maintain a vertical (0$^\circ$) end-effector orientation while performing forward-reach task. Right, Top: photos of stroke subject performing experiment without device assistance (left) and with device fully engaged (right). Without device assistance, the subject has some control over end-effector orientation but cannot bring it out of a fully-pronated position. Engaging the cable mechanism both rotates the hand into a more vertical orientation and allows the subject to better maintain this angle.}
    \label{fig:angles}
    \vspace{-0.4cm}
\end{figure*}

Fig. \ref{fig:mckiesplint} shows a qualitative comparison between reaching forward while wearing a commercial elastic supinator band (McKie Supination Strap and thumb splint) and while wearing our device. Although we do not assist shoulder rotation, we found the resulting end-effector angles to be visually similar to those from a device that  anchors around the bicep.

Fig. \ref{fig:bottle} shows a simulated real-world task of grasping and maneuvering a bottle to qualitatively evaluate ease of reach-to-grasp along with smoothness of arm motion. Without device assistance, the subject could only grasp the top of the bottle from a pronated position. By using the device, the subject could grasp the bottle from the side and could lift it to their face as if to drink.

We quantitatively evaluated device assistance by measuring end-effector angle while the subject attempted to reach forward while maintaining the inclinometer in a vertical orientation. As with the elbow-bent assessment, this test began with the subject's arm placed on the armrest with elbow bent and tucked next to the body. The subject was instructed to slowly lift their arm and reach forward, extending the elbow. When the arm was fully straightened, the subject attempted to hold the inclinometer as vertical as possible for a 30-second duration. Angle results for the straightened-arm portion of the exercise are shown in \mbox{Fig. \ref{fig:angles}.}  

Fig. \ref{fig:angles} shows the maximum and minimum extents of inclinometer readings recorded when the subject attempted to keep the hand in a vertical orientation after reaching forward. Although the subject's arm could be passively rotated into supination, the subject lacked the volitional ability to rotate out of a horizontal hand pose. Overall, most angle deviations occured from the arm moving in space as the subject exerted effort to both rotate their hand and hold their arm in the air. When wearing the device, the rigid brace provides a slight bias towards vertical hand orientation; although the range of angles between \mbox{\textit{No Device}} and \textit{Device Unlocked} are similar, the latter is shifted about 15$^\circ$ towards neutral. Fully engaging the device further rotated the hand, achieving a total PS ROM gain of about $60^\circ$. The locking mechanism also reduced effort required to maintain this position. The range of end-effector angles recorded with the device fully-engaged is narrower than with the other conditions, partially because cable tension resisted forearm pronation and partially because the elbow shook less during the attempt. The subject provided feedback that they felt the exoskeleton made performing this task easier; one possibility is that the device assistance allowed the subject to only exert the minimal effort required to lift the arm, minimizing spasticity.

These results are only for a single subject, but indicate that for this individual, the exoskeleton was comfortable and able to effectively supinate the forearm in order to achieve and maintain a near-neutral hand position. For this individual, our device improved PS ROM by about $60^\circ$ and visually matches performance of a commercial passive splint, all while retaining a below-elbow footprint. Although we do not achieve a truly neutral hand orientation, we do find that reorienting the hand to a $15$--$30^\circ$ pose provided a noticeable assistive impact for some functional tasks. Our findings are congruent with those reported by Chen and \mbox{Lum \cite{chenj2018},} who also found that even an incomplete PS ROM gain \mbox{($38.8\pm 32.2^\circ$)} was sufficient to aid functional tasks.

\section{Conclusions}

In this paper, we present a novel cable-based exoskeleton to assist supination for hemiparetic stroke. We demonstrate feasibility of an approach that leaves the elbow unencumbered; characterization tests suggest that the device can withstand the torques required to adjust forearm rotation even given soft-tissue compliance. We successfully integrate our proposed supination exoskeleton with an existing robotic orthosis and demonstrate that our device does not interfere with hand-opening. Preliminary testing on a stroke subject with minimal volitional supination suggests functional utility. We plan to explore simultaneous actuated strategies for hand-opening and forearm rotation in future work.

Positioning the forearm in a more neutral position enables stroke survivors to approach vertically oriented objects with a more functional hand position during forward reach. This has the potential to increase ease of performance during common bimanual tasks such as grasping a bottle to open it, holding a cup when pouring, or opening the refrigerator to retrieve food. Although internal rotation of the shoulder also contributes to downward orientation of the palm during forward reach for stroke survivors, this device has demonstrated the potential to address part of the problem.


\end{document}